\documentclass[10pt,twocolumn,letterpaper]{article}

\PassOptionsToPackage{table}{xcolor} 

\usepackage{iccv}              %

\definecolor{iccvblue}{rgb}{0.21,0.49,0.74}
\usepackage[pagebackref,breaklinks,colorlinks,allcolors=iccvblue]{hyperref}

\usepackage{enumitem}
\usepackage{booktabs} %
\usepackage{siunitx}
\usepackage{nicefrac}
\robustify\cellcolor
\newcommand{\bgl}{\cellcolor[HTML]{DDDDDD}}
\newcommand{\bgd}{\cellcolor[HTML]{BBBBBB}}
\newcolumntype{Z}{>{\setbox0=\hbox\bgroup}c<{\egroup}@{\hspace*{-\tabcolsep}}}
\newcommand{\xmark}{\sffamily X}%

\title{Spectral Image Tokenizer}

\author{Carlos Esteves \and Mohammed Suhail \\
{\tt\small \{machc,suhailmhd,makadia\}@google.com} \\
Google Research\\
\and Ameesh Makadia
}

\begin{document}
\maketitle

\begin{abstract}
Image tokenizers map images to sequences of discrete tokens,
and are a crucial component of autoregressive transformer-based image generation.
The tokens are typically associated with spatial locations
in the input image, arranged in raster scan order, which is not ideal for
autoregressive modeling. 
In this paper, we propose to tokenize the image spectrum instead, 
obtained from a discrete wavelet transform (DWT), such
that the sequence of tokens represents the image in a coarse-to-fine fashion.
Our tokenizer brings several advantages:
1) it leverages that natural images are more compressible at high frequencies, 
2) it can take and reconstruct images of different resolutions without retraining,
3) it improves the conditioning for next-token prediction --
instead of conditioning on a partial line-by-line reconstruction of the image,
it takes a coarse reconstruction of the full image,
4) it enables partial decoding where the first few generated tokens can reconstruct a
coarse version of the image,
5) it enables autoregressive models to be used for image upsampling.
We evaluate the tokenizer reconstruction metrics as well as
multiscale image generation, text-guided image upsampling and editing.
\end{abstract}

\section{Introduction}
\label{sec:intro}

\begin{figure*}[t!]
  \centering
   \includegraphics[width=\linewidth]{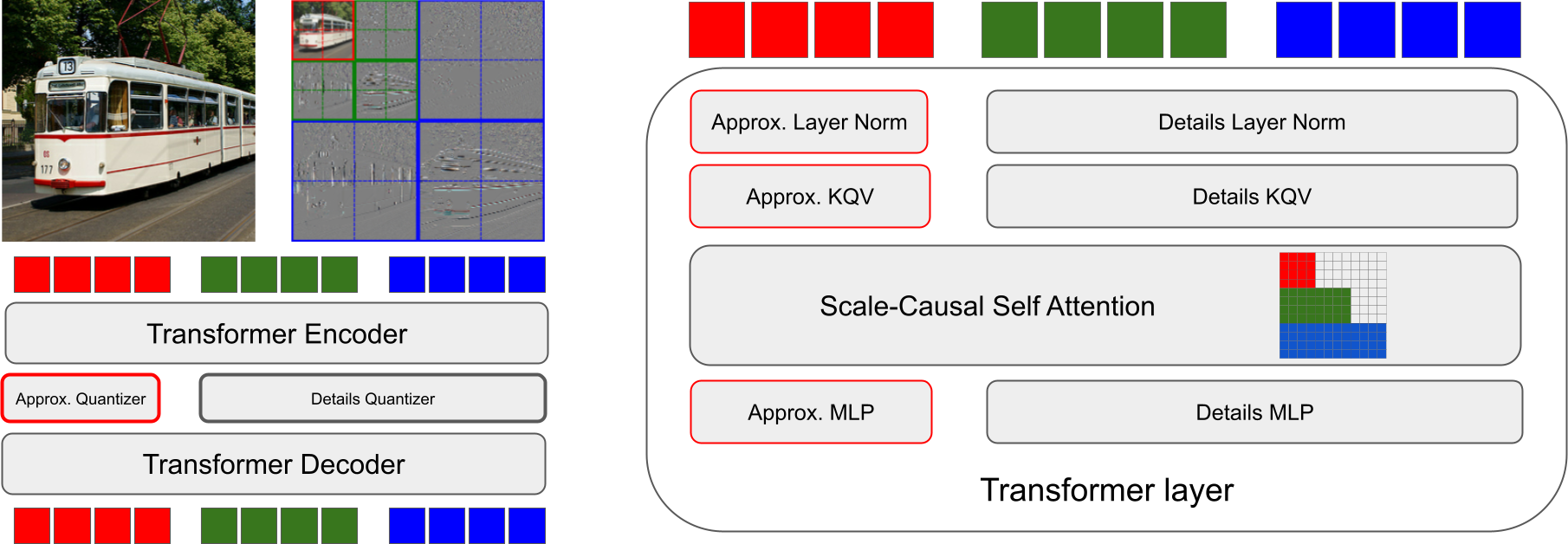}
   \caption{\textit{Left:} we introduce a Spectral Image Tokenizer (SIT),
     that learns to encode and decode discrete wavelet transform (DWT) coefficients to
     and from a small set of discrete tokens, such that the sequence
     represents the image in a coarse-fine fashion. SIT is naturally multiscale
     and enables coarse-to-fine autoregressive image generation with our AR-SIT model.
     SIT also leverages the sparsity of high frequency coefficients in natural images.
     \textit{Right:} details of the encoder/decoder transformer layers. The main architectural
     difference with respect to previous tokenizers is that the distributions of DWT approximation
     and details coefficients are distinct, hence we use specialized parameters for each in
     the quantizer codebooks and inner transformer layers. We also introduce a scale-causal attention
     where each token attends to its own scale and lower scales, which enables encoding, decoding,
     generating, and upsampling images at different resolutions.
     \vspace{-3ex}
   }
   \label{fig:model}
\end{figure*}

In natural language processing, tokenization associates sets of characters to entries
in a vocabulary, which is straightforward since language is discrete.
The sequence of tokens is suitably modeled as a categorical distribution
with autoregressive (AR) transformers, which is the foundation of modern large language models
(LLMs)~\cite{openai2023gpt4,team2023gemini,dubey2024llama}.

While natural images are represented by discrete pixel values,
they exhibit high dimensionality, redundancies, and noise so it's impractical to associate
one token per pixel. This motivated a long line of research of learnable
image tokenizers~\cite{vqvae,vqvae2,vqvae2,vqgan,yu2022vectorquantized}.
While there are successful autoregressive image
generation models~\cite{vqgan,yu2022vectorquantized,yu2022scaling}, 
images are not sequential like language,
which motivated developing alternatives such as
denoising diffusion models~\cite{pmlr-v37-sohl-dickstein15,diffusionbeatsgans,latentdiffusion}
and masked transformers~\cite{chang2022maskgit,pmlr-v202-chang23b,titok}.
Nevertheless, most of these alternatives also operate on the latent space of tokenizers
like VQGAN~\cite{vqgan} instead of raw pixels.

In this work, we revisit AR transformer-based image generation. Our main contribution is a tokenizer operating on the image spectrum, specifically on DWT coefficients,
where the coarse-to-fine representation lends itself more naturally to a sequential interpretation.
Our Spectral Image Tokenizer (SIT) has several useful properties and enables different applications:

\begin{enumerate}[label=P.\arabic*]
\item \label{prop:spectrum}
  Since the power spectrum of natural images decreases with frequency, high frequencies can be
more heavily compressed with little effect in visual quality. SIT leverages this by
associating tokens to larger patches at higher wavelet scales than at lower scales (see \cref{fig:patches}).
\item \label{prop:tok_multiscale}
  SIT is transformer-based~\cite{transformer};
by using an attention mask where each scale depends
on itself and lower scales (``Scale-Causal attention''), SIT can be trained
at a single resolution and used to tokenize
images of multiple resolutions (any number of scales up to the trained maximum),
and detokenize partial token sequences (up to some scale), reconstructing a coarse image.
\item \label{prop:ar_conditioning}
  Using SIT, we train an autoregressive generative transformer (AR-SIT) that
models images coarse-to-fine. The next-token prediction is then conditioned on a coarse
reconstruction of the image given by the partial token sequence,
instead of the usual conditioning on the partial reconstruction of the previous rows of the image.
\item \label{prop:ar_multiscale}
  AR-SIT can quickly generate only the first few tokens and reconstruct a
coarse version of the image, enabling applications like quickly showing multiple coarse generations
and letting the user select which ones to refine.
\item \label{prop:upsampling}
  AR-SIT can be used for text-based upsampling of an input low resolution image,
by starting the decoding process with the few tokens output by SIT, and generating the
rest of the sequence up to a desired resolution.
\item \label{prop:editing}
  AR-SIT can be used for text-guided image editing,
by encoding a given image up to a coarse scale, and generating the finer details
conditioned on a new caption.
\end{enumerate}

Currently, image generation is dominated by diffusion models such as Imagen 3~\cite{imagen3},
DALL-E-3~\cite{BetkerImprovingIG} and Stable Diffusion 3~\cite{stablediffusion3}.
On the other hand, LLMs such as
Gemini~\cite{team2023gemini}, GPT-4~\cite{openai2023gpt4}, and Llama 3 \cite{dubey2024llama}
are based on autoregressive transformers.
We believe autoregressive image generation is still worth pursuing because it might benefit
from advances in LLMs, and multimodal applications might benefit of having a single architecture
for all modalities.
Recent work on image and video generation support this point~\cite{yu2024language,VAR,sun2024autoregressive}.
\citet{dieleman2024spectral} recently interpreted denoising diffusion models as
spectral autoregression, since, when looking at image spectra, the denoising procedure
uncovers frequencies from low to high. In contrast, our method does literal spectral
autoregression. 

\section{Related work}
\label{sec:related}
\paragraph{Image tokenization}
Several methods have been developed to map images to a small set of discrete tokens
suitable for generative modeling.
VQ-VAE~\cite{vqvae} introduced vector-quantization in the latent space of a Variational Auto-Encoder to map images, audio and video to a set of discrete values.
VQGAN~\cite{vqgan} improved upon VQVAE by using perceptual and adversarial losses.
We build on ViT-VQGAN~\cite{yu2022vectorquantized}, which improved upon VQGAN by using a
Vision Transformer (ViT)~\cite{dosovitskiy2021an} instead of convolutions,
as well as codebook factorization and feature normalization.

In this paper, we are interested in multiscale image representations.
VQ-VAE-2~\cite{vqvae2} introduced multiscale latents by keeping and quantizing intermediate
downsampled convolutional features.
RQ-VAE~\cite{rqvae} quantized a set of residuals such that the latent vector is
represented in a coarse-to-fine fashion and reconstructed by adding the code embeddings
for each residual.
Similarly, VAR~\cite{VAR} and STAR~\cite{ma2024starscalewisetexttoimagegeneration}
split the latent space in multi-scale quantized residuals,
so that it can be reconstructed by upsampling and summing.
FAR~\cite{yu2025frequencyautoregressiveimagegeneration} splits the latents into frequencies to
train a next-frequency generative model. 
DQ-VAE \cite{dqvae} encoded smooth regions with fewer tokens that are generated first,
while QG-VAE~\cite{elsner2024quantisedglobalautoencoderholistic}
and SEMANTICIST~\cite{wen2025principalcomponentsenablenew}
learn non-local tokens that are ordered by importance.

The crucial difference between our approach and the aforementioned is that
we operate on spectral coefficients of the input and not on latent features.
This allows our tokenizer to be truly multiscale
so that it can take inputs at different scales and reconstruct up to some scale,
while also leveraging that higher frequencies are more compressible in natural images.

Tangentially related to our work,
Wave-ViT~\cite{wavevit2022} modified the ViT self-attention by
applying a DWT to the input and concatenating coefficients,
effectively exchanging space for depth to reduce the sequence length. 
\citet{zhu2024waveletbasedimagetokenizervision} modified the ViT patchifier
in a similar way, and introduced patch embeddings that leverage the sparsity of high frequency
coefficients.
Both methods are for discriminative tasks such as image classification and segmentation,
while we focus on generation.

\vspace{-0.5cm}
\paragraph{Autoregressive image generation}
Early approaches PixelRNN and PixelCNN~\cite{pixelrnn,NIPS2016_b1301141} generate
images pixel by pixel by modeling the conditional distribution of the pixel given the previous pixels
with recurrent layers of causal convolutions.
PixelSNAIL~\cite{pmlr-v80-chen18h} improved on this model by introducing self-attention layer
to better model long-range dependencies.
VQ-VAE~\cite{vqvae} introduced a two-stage approach with a tokenizer and a separate stage
to model the distribution of tokens.
VQGAN~\cite{vqgan} greatly improved these results by both improving the tokenizer
and using a transformer to model the distribution.
Finally, ViT-VQGAN~\cite{yu2022vectorquantized} proposed a transformer-based tokenizer,
which Parti~\cite{yu2022scaling} used with a large autoregressive transformer capable
of high-quality text-to-image generation.

MaskGIT~\cite{chang2022maskgit} and Muse~\cite{pmlr-v202-chang23b} highlighted the disadvantages of
the typical autoregressive models raster-order conditioning, and proposed to generate all tokens
in parallel iteratively, where each iteration keeps the highest confidence tokens.
We address the same problem with a tokenizer whose sequence represents
the image in a coarse-to-fine order instead of raster-order.

\vspace{-0.5cm}
\paragraph{Multiscale image generation}
Multiscale image generation ideas have appeared in the context of VAEs~\cite{Cai_2019},
GANs~\cite{karras2018progressive,rottshaham2019singan}, and diffusion models~\cite{gu2024matryoshka,kulikov2023sinddm,ryu2022pyramidal},
but have not been sufficiently explored with AR transformers.

\citet{abs-2103-03841} represented an image as a sequence of quantized and thresholded DCT
coefficients, where the compression comes from the fact that many coefficients are zero
and are omitted. By sorting the sequence by frequency,
the method can model images from coarse-to-fine like we do.
However, the compressed representation is handcrafted and results in long sequences.
In a similar vein, \citet{mattar2024waveletsneedautoregressiveimage} handcrafted a tokenizer
based on DWT coefficients, introducing tokens to represent large chunks of zeros. It also
results in long sequences and is only applied to generations of small grayscale images.
In contrast with these methods, instead of handcrafting a compressed input representation,
our tokenizer learns to encode to, and decode from, a short compressed coarse-to-fine sequence.

\section{Background}
A discrete wavelet transform (DWT) is based on successive convolutions of
the signal $f$ with a pair of lowpass $g$ and highpass $h$ filters with the first step as follows
\begin{align}
  f_{\text{low}_1}[n] &= f \star g = \sum_k f[k]g[n-k], \\
  f_{\text{high}_1}[n] &= f \star h = \sum_k f[k]h[n-k].
\end{align}
The high and low outputs are subsampled by a factor of two,
and the operation is repeated for the low channel, such
that at level $L$ we compute $f_{\text{low}_L} = f_{\text{low}_{L-1}}\star g$
and $f_{\text{high}_L} = f_{\text{low}_{L-1}}\star h$, subsample again and drop $f_{\text{low}_{L-1}}$.
The output at level $L$ comprises the approximation coefficients $f_{\text{low}_L}$
and the detail coefficients $\{f_{\text{high}_k}\}$ for ${1 \le k \le L}$.
This output has the same cardinality as the input and the transformation is invertible,
where the forward transform is typically called ``analysis'' and the backward ``synthesis''.
The simplest wavelet family is the Haar, where $g = [1, 1]^\top$ and $h = [1, -1]^\top$ (optionally
scaled to unit norm).

\begin{figure}[t]
  \centering
   \includegraphics[width=\linewidth]{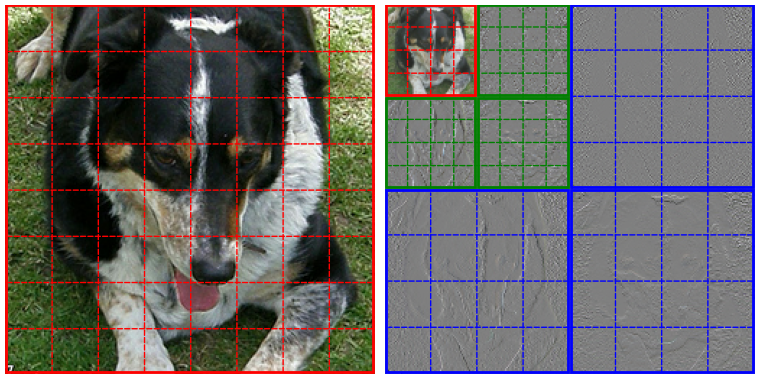}
   \caption{Input patchification. \textit{Left:} typical patchification for Vision Transformers (ViT)~\cite{dosovitskiy2021an}, where the image is split in equal-sized patches.
     \textit{Right:} we propose to patchify the coefficients of a
     discrete wavelet transform (DWT) instead. Each scale is shown in a different color.
     Scales other than the lowest contain three blocks representing horizontal,
     vertical and diagonal details;
     we concatenate the spatially correspondent patches of each block such that each scale
     is represented by the same sequence length.
     This results in larger patch sizes for higher scales,
     which are more compressible.
     The figure shows 3 scales and 16 tokens per scale; in our experiments we use 4 or 5 scales
     and 256 tokens per scale.
     \vspace{-3ex}
   }
   \label{fig:patches}
 \end{figure}

For image analysis we use a 2D DWT, which is obtained by simply convolving the rows and
columns with $g$ and $h$. The approximation coefficients $f_{\text{low}_1} = f\star (gg^\top)$, and
the details are divided into horizontal $f_{\text{H}_1} = f\star (g h^\top)$,
vertical $f_{\text{V}_1} = f\star (h g^\top)$ and diagonal $f_{\text{D}_1} = f\star (h h^\top)$ details.
Subsequent levels apply the same operations to the approximation coefficients.
\cref{fig:patches} shows a two-level transform,
where we can see that the approximation coefficients correspond to a coarse version of the image.

We refer to the textbooks by \citet{mallatbook} and \citet{daubechiesbook} for more information
about wavelets.

\section{Method}
\label{sec:method}
Our main contribution, SIT, is an image tokenizer
that takes discrete wavelet transform (DWT) coefficients.
The model follows ViT-VQGAN~\cite{yu2022vectorquantized}, with important changes that we describe
in this section and visualize in \cref{fig:model}. For image generation, we introduce AR-SIT which is based on Parti~\cite{yu2022scaling} with minor changes, using SIT as the tokenizer.

\subsection{Tokenizer}
\label{sec:tokenizer}
\paragraph{Patchification and featurization}
The first step is to map the input to a sequence of patches.
We apply the Haar DWT on the input image and patchify each scale separately.
While Haar is the simplest wavelet and lacks properties found in other wavelets
useful for compression,
we found no benefits of using other wavelet families such as CDF~\cite{cdf}.

Our design choice is to use the same number of patches for each scale. In a DWT, the higher scales
correspond to high frequency details which are represented by more coefficients than the
lower scales, but contribute less to the spatial pixel values.
In other words, in natural images, most of the power spectrum is concentrated on lower frequencies.
By representing each scale with the same number of tokens, we are compressing more the higher
frequencies (since they are represented by more coefficients),
similarly to what is done in image compression methods such as JPEG2000~\cite{jpeg2000}.

The approximation (or lowpass) DWT coefficients correspond to a coarse version of the input image,
which we consider the first scale.
The following scales are divided in three blocks corresponding to horizontal, vertical, and diagonal
details, where each coefficient relates to a specific spatial location. Thus,
we can concatenate the three blocks such that each entry corresponds to the same
spatial location.
For example, the first scale will typically be split in patches that are $32\times32\times3$, with
channels corresponding to RGB, the second scale will be $32\times32\times9$, where the channels correspond
to the RGB of horizontal, vertical, and diagonal details, the third will be $64\times64\times9$ and so on.
\cref{fig:patches} shows an example of our patchification scheme.

Since patches have different resolutions (higher scales will have larger patches),
the usual ViT linear embedding to map patches to features cannot be shared across all patches
so we have different parameters per scale.
Formally, given an image $f$, our patchifiers compute
\begin{align}
  & f_{\text{low}_L},\, \{f_{\text{H}_i}\}_{i \le L},\,\{f_{\text{V}_i}\}_{i \le L},\,\{f_{\text{D}_i}\}_{i \le L} = \text{DWT}(f), \\
  & c_1 = P_0(f_{\text{low}_L}), \\
  & c_s = P_s(f_{\text{H}_{L-s+2}},\, f_{\text{V}_{L-s+2}},\, f_{\text{D}_{L-s+2}}),\; 1 < s \le L,
\end{align}
where $c_s = \{c_s^n\}_{1 \le n \le N}$ is the sequence of token embeddings at scale $s$,
and $c_s^n \in \mathbb{R}^C$ is the embedding of the $n$-th token at the $s$-th scale, 
while $L$ is the number of DWT levels,
$S=L+1$ is the number of scales and $N$ the number of tokens per scale.
For brevity, when there is no ambiguity,
we may omit the set indexing and use $\{c_s\}$ to denote the set of tokens of all
scales, for example.

The final projection after the decoder needs to map the features back to different
sized patches, so it also has different parameters per scale.
Those patches still represent DWT coefficients so an inverse DWT (IDWT) is finally applied
to obtain an image output.

\vspace{-0.5cm}
\paragraph{Flexible sequence length}
The tokenizer encoder and decoder transformers operate on the sequence
of patch features of length $SN$.
The sequence length is a major factor in resource utilization so we want to keep it constrained. 
Our method is flexible since we can choose the number of scales and the patch size per scale, while
most ViT-based models such as ViT-VQGAN~\cite{yu2022vectorquantized} are more restricted.
They use the same patch size for the whole image; thus, keeping the same patch size and doubling each
image dimension would increase the sequence length by a factor of 4,
where our method is capable of including additional scales which only
increases the sequence length by multiples of $N$.

For example, for a $256\times 256$ input, ViT-VQGAN uses $8\times 8$ patches to obtain a
sequence of 1024 elements.
Our SIT can use 4 scales and 256 tokens per scale for the same sequence length.
When increasing the resolution to $512\times 512$, the baseline can either increase the patch size to
$16\times 16$, resulting in the same sequence length, or keep it $8\times 8$, resulting in a 4$\times$ longer sequence. SIT, for example, can vary the number of scales from 4 to 6, resulting in
sequence lengths of 1024, 1280, 1536 which are more manageable. 

\vspace{-0.5cm}
\paragraph{Transformers}
After featurization, the single sequence containing all scales
passes through a transformer encoder, followed by quantization and a transformer decoder.
We propose two optional modifications to the transformers, which are otherwise identical to
the ones used in ViT~\cite{dosovitskiy2021an}.

First, we propose a scale-causal attention mask, where an element at some scale attends to
all elements of its own scale and lower scales, represented by a block-triangular mask pattern.
With dense attention, we write the application of the encoding transformer
$T_\text{enc}$ as $\{z_s\} = T_\text{enc}(\{c_s\})$,
and for scale-causal we write
$z_s = T_\text{enc}(\{c_k\}_{1 \le k \le s})$ for each $s$.
The scale-causal attention can be applied independently to the encoder and decoder,
enabling different applications.
For the multiscale reconstruction experiments in \cref{sec:multiscale_reconstruction}, we need
to both encode and decode multiple resolutions, so both encoder and decoder use scale-causal masks.
For coarse-to-fine image generation in \cref{sec:partial_generation}, only the decoder needs to
be scale-causal in order to decode the partially generated sequence.
For the text-guided image upsampling in \cref{sec:upsampling},
only the encoder needs to be scale-causal to encode the lower resolution inputs.
For the image editing experiments in \cref{sec:editing},
the scale-causal encoder prevents information leaking from high to low scale.

Second, we propose to use different transformer parameters for the
approximation (first scale) and details coefficients (other scales).
In our model, the subsequences corresponding to coefficients of each type
of coefficient come from quite distinct distributions, so it makes sense to treat them differently.
This contrasts with the spatial representation of images where each patch can be considered
as coming from the same distribution.
Thus, the parameters of the key, query, and value embeddings,
the layer norms, and the MLP on each transformer layer are not shared between the
approximation and details coefficients.
This Approximation-Details Transformer (ADTransformer)
still takes a single sequence composed of all coefficients.
We experimented with different transformers per sequence
with cross-attention for information sharing,
but it performed worse.
This change leads to more memory utilization to store the extra parameters,
but the training/inference speed is similar because the number of operations is unchanged. 

\begin{table}[t]
  \begin{center}
    \resizebox{\columnwidth}{!}{
      \begin{tabular}{@{}lc
        S[table-format=0.3,table-auto-round]
        S[table-format=2.1,table-auto-round]
        S[table-format=0.3,table-auto-round]
        S[table-format=1.2,table-auto-round]
        S[table-format=3.1,table-auto-round]
        S[table-format=3.0,table-auto-round]
        }
      \toprule
                                                 &  & {LPIPS $\downarrow$} & {PSNR $\uparrow$}  & {L1 $\downarrow$}    & {FID $\downarrow$} & {IS $\uparrow$}    & {images/s $\uparrow$} \\[0pt]
      \midrule
        \textit{Resolution:} $\mathit{512\times 512}$ &  &             &             &             &           &             &                \\[0pt]
      ViT-VQGAN                                  &  & 0.3196      & \bgl 22.435 & \bgl 0.0415 & 6.915     & 151.5       & \bgd 592.8     \\
      SIT-5 (Ours)                               &  & \bgl 0.2600 & 22.01       & 0.0457      & \bgl 2.65 & \bgl 191.97 & \bgl 410.34    \\
      SIT-6 (Ours)                               &  & \bgd 0.2394 & \bgd 23.06  & \bgd 0.0404 & \bgd 1.74 & \bgd 203.70 & 319.89         \\
      \midrule
      \textit{Resolution:} $\mathit{256\times256}$    &  &             &             &             &           &             &                \\[0pt]
      ViT-VQGAN (reported)                       &  & {-}         & 24.76       & 0.0322      & 1.99      & 184.4       & {-}            \\
      ViT-VQGAN (reproduced)                     &  & 0.1665      & \bgd 24.95  & \bgd 0.0309 & 2.33      & 183.99      & {-}            \\
      ViT-VQGAN (no LL)                          &  & 0.1626      & 23.76       & 0.0382      & \bgl 1.20 & 194.58      & \bgd 626.40    \\
      SIT-4 (Ours)                               &  & \bgl 0.1435 & 24.01       & 0.0371      & \bgl 1.20 & \bgl 199.53 & \bgl 596.36    \\
      SIT-5 (Ours)                               &  & \bgd 0.1353 & \bgl 24.48  & \bgl 0.0350 & \bgd 0.97 & \bgd 202.29 & 411.35        \\
      SIT-SC-5 (Ours)                            &  & 0.1608      & 24.12       & 0.0366      & 1.33      & 193.70      & 411.35        \\
      \midrule
      \textit{Resolution:} $\mathit{128\times128}$    &  &             &             &             &           &             &                \\[0pt]      
      ViT-VQGAN                                  &  & 0.1845      & 26.26       & 0.0298      & 3.77      & 117.28      & \bgd 626.15    \\
      SIT-SC-5 (ours)                            &  & \bgd 0.1585 & \bgd 27.11  & \bgd 0.0266 & \bgd 2.13 & \bgd 129.27 & 582.05         \\
      \midrule
      \textit{Resolution:} $\mathit{64\times64}$      &  &             &             &             &           &             &                \\[0pt]
      ViT-VQGAN                                  &  & 0.1292      & 28.78       & 0.0227      & 3.53      & 20.96       & 626.87         \\
      SIT-SC-5 (ours)                            &  & \bgd 0.1107 & \bgd 31.29  & \bgd 0.0173 & \bgd 1.39 & \bgd 30.13  & \bgd 847.32    \\
      \midrule
      \textit{Resolution:} $\mathit{32\times32}$      &  &             &             &             &           &             &                \\[0pt]            
      ViT-VQGAN                                  &  & 0.2142      & 23.28       & 0.0454      & {-}       & \bgd 3.70   & 626.62         \\
      SIT-SC-5 (ours)                            &  & \bgd 0.0294 & \bgd 36.83  & \bgd 0.0098 & \bgd 0.31 & 3.52        & \bgd 824.79    \\
      \midrule
      \textit{Resolution:} $\mathit{16\times16}$      &  &             &             &             &           &             &                \\[0pt]            
      ViT-VQGAN                                  &  & 0.1268      & 24.93       & 0.0388      & {-}       & 1.66        & 626.67         \\
      SIT-SC-5 (ours)                            &  & \bgd 0.0129 & \bgd 41.25  & \bgd 0.0063 & \bgd 0.09 & \bgd 1.81   & \bgd 2620.09   \\
      \bottomrule
    \end{tabular}
    }
  \end{center}
  \caption{Multiscale reconstruction on ImageNet.
    ``SC'' denotes scale-causal attention, which slightly reduces performance at the highest
    resolution but enables multiscale reconstruction without downsampling/upsampling or retraining.
    The ViT-VQGAN values from~\citet{yu2022vectorquantized} used a logit-laplace loss (LL)
    which was later considered harmful~\cite{yu2022scaling}, so we retrain without it.
    Our SIT improves reconstruction metrics, is significantly faster at lower resolutions,
    and robust when increasing resolution.
    The ViT-VQGAN baseline suffered from instability during training on $512\times 512$ inputs so we selected the
    best values before divergence.
    We report test time throughput of an encoding/decoding cycle for the max batch size
    that fits on a TPU v5e.
    }
  \label{tab:multiscale_reconstruction}  
\end{table}

\begin{table}[t]
  \begin{center}
    \resizebox{\columnwidth}{!}{
      \begin{tabular}{@{}lc
        S[table-format=2.1,table-auto-round]
        S[table-format=2.1,table-auto-round]
        S[table-format=2.1,table-auto-round]
        S[table-format=1.1,table-auto-round]
        }      
        \toprule
                                                &  & {FID $\downarrow$}  & {IS $\uparrow$}   & {images/s $\uparrow$} & {images/Gb $\uparrow$} \\[0pt]
        \midrule
        \textit{Resolution:} $\mathit{256\times256}$ &  &            &            &                &                 \\[0pt]
        Parti350M (reported)                    &  & 14.10      & {-}        & {-}            & {-}             \\[0pt]
        Parti350M                               &  & \bgd 12.35 & 36.47      & \bgd 7.75      & \bgd 12         \\
        AR-SIT-SCD-4 (Ours)                     &  & 12.57      & \bgd 37.28 & 6.50           & 8               \\
        \midrule
        \textit{Resolution:} $\mathit{128\times128}$ &  &            &            &                &                 \\[0pt]
        Parti350M                               &  & \bgd 11.19 & \bgd 33.49 & 7.57           & \bgd 12         \\
        AR-SIT-SCD-4 (Ours)                     &  & 11.42      & 33.24      & \bgd 12.59     & \bgd 12         \\
        \midrule
        \textit{Resolution:} $\mathit{64\times64}$   &  &            &            &                &                 \\[0pt]
        Parti350M                               &  & \bgd 10.51 & 16.91      & 7.61           & 12              \\
        AR-SIT-SCD-4 (Ours)                     &  & 11.36      & \bgd 18.60 & \bgd 24.52     & \bgd 16         \\
        \midrule
        \textit{Resolution:} $\mathit{32\times32}$   &  &            &            &                &                 \\[0pt]
        Parti350M                               &  & \bgd 5.81  & 2.90       & 7.66           & 7.66            \\
        AR-SIT-SCD-4 (Ours)                     &  & 7.60       & \bgd 3.17  & \bgd 74.66     & \bgd 28         \\
        \bottomrule
      \end{tabular}
    }
  \end{center}
  \caption{Coarse-to-fine image generation on MS-COCO~\cite{LinMBHPRDZ14}.
    With SIT, the autoregressive generation is stopped early for a coarse
    version of the image. The Parti350M baseline does not have this property, so
    we generate at full resolution and downsample for comparison.
    AR-SIT matches the baseline performance at the training resolution
    but is several times faster and more memory efficient 
    at lower resolutions,  even when trained only on higher resolution data.
    We report throughput and memory utilization during generation
    given the max batch size that fits on a TPU v6e.
  }
  \label{tab:partial_generation}
\end{table}

\vspace{-0.5cm}
\paragraph{Quantizer}
The encoder outputs a sequence of features that are quantized to a fixed-sized codebook,
similarly to ViT-VQGAN~\cite{yu2022vectorquantized}, VQVAE~\cite{vqvae}, and VQGAN~\cite{vqgan}.
We modify the quantizer such that approximation and details have different codebooks for the same
reasons discussed previously.
Thus, codebook sizes and feature dimensions are the same as the ViT-VQGAN baseline,
but we have different features for the same code at different positions.
Formally, we apply $q_s^n = Q_{\text{approx}}(z_s^n)$ for $s=1$ and 
$q_s^n = Q_{\text{details}}(z_s^n)$ for $s \ge 2$
for each pair $(s,\, n)$, where $q_s^n$ is chosen from the codebook for $s$
and can be associated with its discrete position in the codebook, denoted $\lfloor q_s^n \rfloor$.

\begin{figure*}[t]
  \centering
   \includegraphics[width=\linewidth]{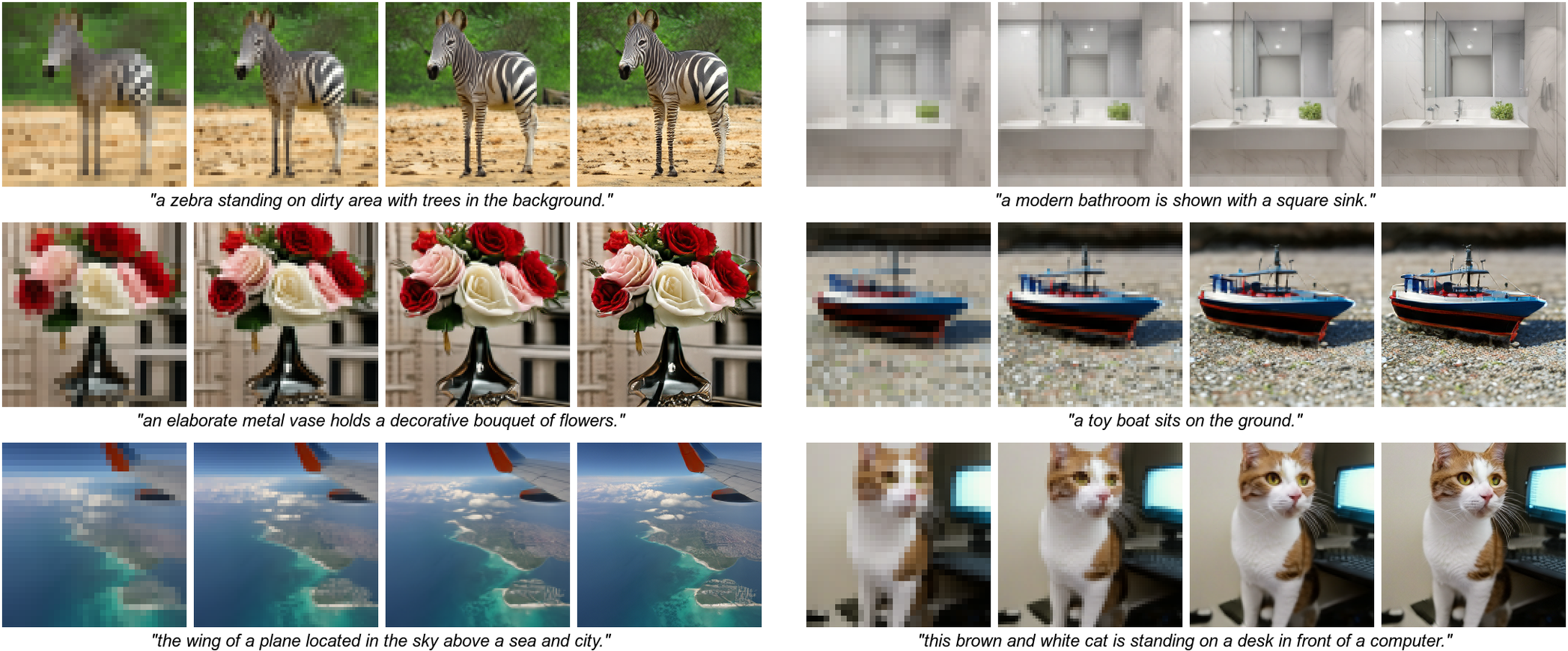}
   \caption{Coarse-to-fine text-to-image on MS-COCO~\cite{LinMBHPRDZ14} prompts.
     Each quadruple shows generations from AR-SIT-SCD for the given prompt, where, from left
     to right, only the first 25\%, 50\%, 75\% and 100\% of tokens are generated,
     corresponding to resolutions of $32\times32$ to $256\times256$, where only
     $256\times 256$ images are seen during training.
     Our model enables quick generation of coarse image candidates that can be further
     improved if needed.
     \vspace{-0.5cm}
   }
   \label{fig:partial_generation}
\end{figure*}

\vspace{-0.6cm}
\paragraph{Training}
We follow the ViT-VQGAN training protocol and use the same weighting for the
L2, perceptual, adversarial, and quantization losses.
We remove the logit-laplace loss that was shown detrimental in follow-up work~\cite{yu2022scaling}.

All losses are applied to the spatial domain images
reconstructed by the inverse discrete wavelet transform on the decoder output:
$\hat{f} = \text{IDWT}(T_{\text{dec}}(\{q_s\}))$.
We noticed instability during training due to the adversarial loss,
which was fixed by applying spectral normalization following~\citet{miyato2018spectral}, which
simply divides the discriminator weight matrices by their largest singular value.

\subsection{Autoregressive image generation}
We use our tokenizer for autoregressive image generation,
by training a second stage transformer model similar to Parti~\cite{yu2022scaling},
with some modifications.
Formally, the autoregressive transformer $T$ models categorical distributions over the discrete
codes
\begin{align}
& P(\lfloor q_s^n \rfloor \mid \{ \lfloor q_i \rfloor \}_{1 \le i < s} \cup \{ \lfloor q_s^i \rfloor \}^{1 \le i < n}) = T(c)
\end{align}
which can be sampled one code at a time for generation. $T$ can be conditioned on a textual description
processed by a transformer encoder for text-to-image generation.
For training, we model the distribution of input codes as
$P(\{\lfloor q_s \rfloor\}) = \prod_{1 \le s \le S}^{1 \le n \le N} P(\lfloor q_s^n \rfloor)$
and minimize the negative log-likelihood $-\log{P(\{\lfloor q_s \rfloor\}) }$ over the training set.

With different codebooks for approximation and details, 
the same token id might have a different meaning depending on its position. 
Thus, the AR model has different token embeddings for approximation and details tokens.
The last layer for logit prediction is also different.

For the generative applications in \cref{sec:partial_generation,sec:upsampling},
we introduce mild changes in order to interrupt the generation after all tokens up to
a certain scale are generated, and to start the generation with given tokens up to
a certain scale.

\section{Experiments}
\label{sec:experiments}

We focus our experiments on demonstrating the properties and
applications enumerated in 
\ref{prop:spectrum}-\ref{prop:editing} (\cref{sec:intro}).
Our ideas build on ViT-based tokenizers, so the most fair comparisons
are against ViT tokenizers and AR generative models; namely 
ViT-VQGAN~\cite{yu2022vectorquantized} and Parti~\cite{yu2022scaling}.
In these comparisons we are able to match the architectures and
training protocol exactly.
Nevertheless we also compare with a broader set of methods in \cref{sec:class_cond}.

\begin{figure*}[t]
  \centering
   \includegraphics[width=\linewidth]{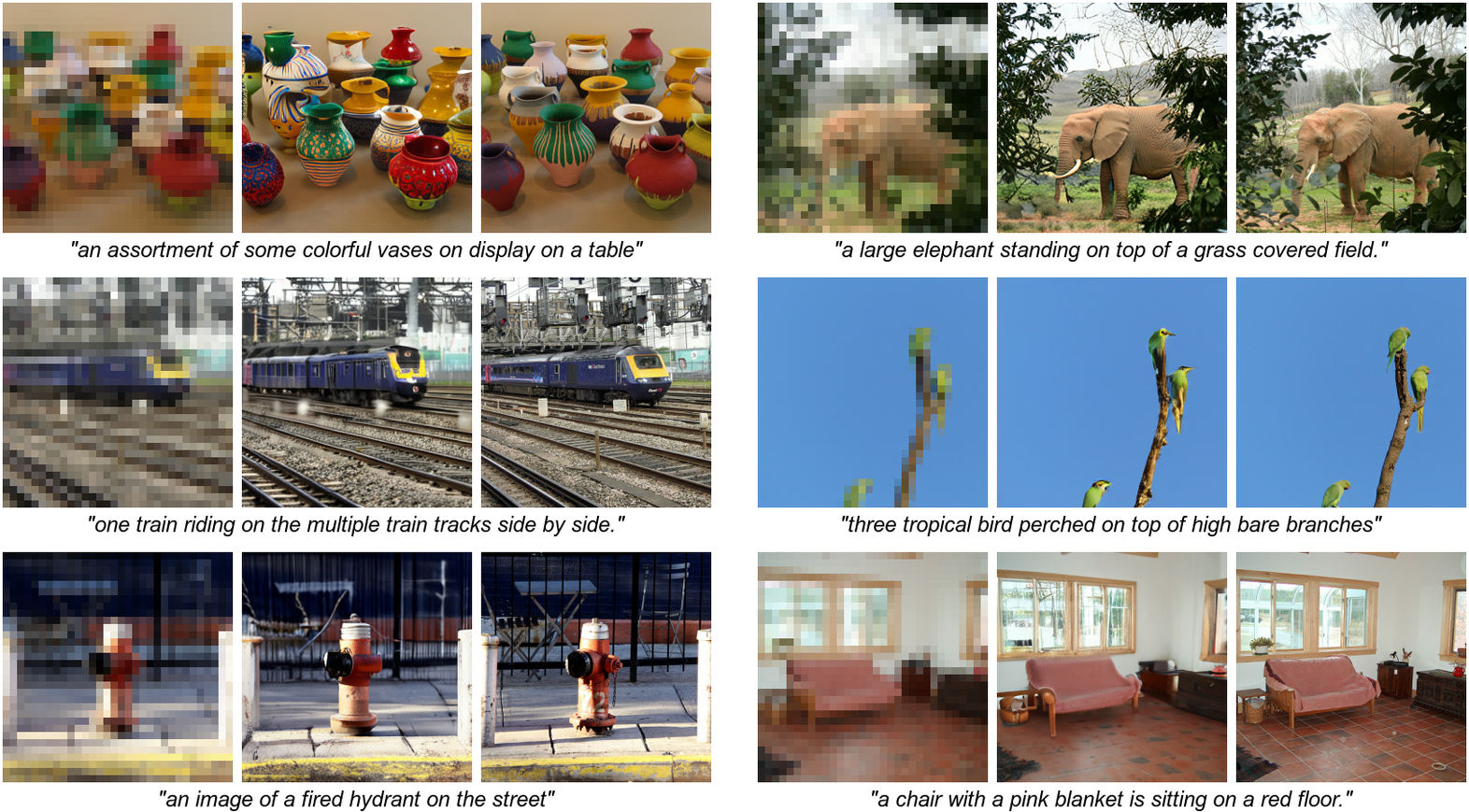}
   \caption{Text-guided image upsampling on MS-COCO~\cite{LinMBHPRDZ14}.
     Our coarse-to-fine generative models can take a low-resolution image, encode it as
     the first few tokens of a sequence, and generate the rest of sequence, which, when decoded,
     effectively upsamples the input.
     Each triplet shows the given $32\times32$ image, our $256\times256$ reconstruction and the ground truth.
     \vspace{-0.5cm}
   }
   \label{fig:superresolution}
 \end{figure*}

\subsection{Multiscale reconstruction}
\label{sec:multiscale_reconstruction}
We train our tokenizer on ImageNet~\cite{deng2009imagenet}
and evaluate its reconstruction performance.
We follow the ViT-VQGAN~\cite{yu2022vectorquantized} ``Small'' architecture
and training protocol.

The model ``SIT-5'' is the Spectral Image Tokenizer with 5 scales
from $16\times16$ to $128\times 128$ resolutions, where each scale is represented
by 256 tokens.
The variation ``SIT-SC'' uses scale-causal attention on both the encoder and decoder,
which enables handling inputs and outputs of different resolutions, even though
it is only trained at $256\times 256$.
For example, when the input image is $64\times64$, only the first three scales are used,
resulting in shorter sequence lengths which reduces memory utilization and processing time.
Both models employ the ADTransformer described in \cref{sec:tokenizer}.

The ViT-VQGAN baseline only works for $256\times256$ inputs, so to evaluate against it fairly,
we upsampled the low-resolution inputs to that resolution. This brings the input patches
away from the training distribution, which might explain the drop in reconstruction quality.

We also evaluate at $512 \times 512$.
The ViT-VQGAN baseline increases its patch size from $8\times 8$ to $16\times 16$,
keeping the sequence length constant.
Interestingly, the baseline suffered heavy instability
during training, which was not resolved by reducing the learning
rate, using spectral normalization, or the logit-laplace loss.
In contrast, our SIT variations trained successfully with no hyperparameter changes.

\cref{tab:multiscale_reconstruction} shows the reconstruction metrics while \cref*{fig:multiscale_reconstruction} (supplemental) shows reconstruction samples at multiple resolutions.
Results provide evidence for \ref{prop:spectrum}
and demonstrate \ref{prop:tok_multiscale} (\cref{sec:intro}).

\subsection{Coarse-to-fine text-to-image generation}
\label{sec:partial_generation}
We tackle text-to-image generation by using an autoregressive transformer
to model the distribution of discrete tokens output by our tokenizer.
Since the SIT sequence of tokens represents an image in a coarse-fine fashion,
the autoregressive model generation has the same property, which means that we
can interrupt the generation after a certain number of tokens and decode a
coarse version of the generation.

For this to work, the SIT decoder must be scale-causal, while there is
no constraint for the encoder so we use dense attention (we denote
this variant Scale-Causal Decoder, or ``SIT-SCD'').
It uses a ``small'' encoder and ``large'' decoder as
described in ViT-VQGAN~\cite{yu2022vectorquantized}, and 4 scales.

We follow the Parti350M~\cite{yu2022scaling} architecture and training
protocol and name our generative model auto-regressive SIT, or ``AR-SIT''.
SIT and AR-SIT  are trained on a subset of
WebLI~\cite{pali} of around 128M images, where
each image is seen at most once during training.
Evaluation is on 30k examples of MS-COCO~\cite{LinMBHPRDZ14}.
We show metrics and generations at different resolutions in \cref{fig:partial_generation} and
\cref{tab:partial_generation}.
AR-SIT matches the baseline performance 
but is several times faster and more memory efficient at lower resolutions.
These results demonstrate the property \ref{prop:ar_multiscale} (\cref{sec:intro}).

\subsection{Text-guided image upsampling}
\label{sec:upsampling}
We leverage the coarse-to-fine nature of SIT to apply a pre-trained
AR-SIT for text-guided image upsampling,
where we are given a low resolution image and a caption.

The idea is to encode the low-resolution image, which will give the first tokens
of our high resolution output. AR-SIT then takes these tokens
and generates the rest of the sequence. 
For this to work, the SIT encoder must be scale-causal to properly
tokenize low-resolution inputs,
while there is no constraint for the decoder so we use dense attention.
We denote this variant Scale-Causal Encoder, or ``SIT-SCE''.
The model is otherwise as the one used in \cref{sec:partial_generation}.

Here the inputs are set to $32\times32$.
\cref{fig:superresolution} shows some text-guided upsampling examples
on MS-COCO~\cite{LinMBHPRDZ14}.  We obtain an FID of 6.2
when evaluating the generations over the whole dataset, compared to
12.6 when given only the prompts.
The supplemental shows additional results upsampling from $16\times 16$ inputs in
\Cref*{fig:upsampling_supp}, and metrics in \cref*{tab:upsampling}.

\begin{figure*}[t]
  \centering
   \includegraphics[width=\linewidth]{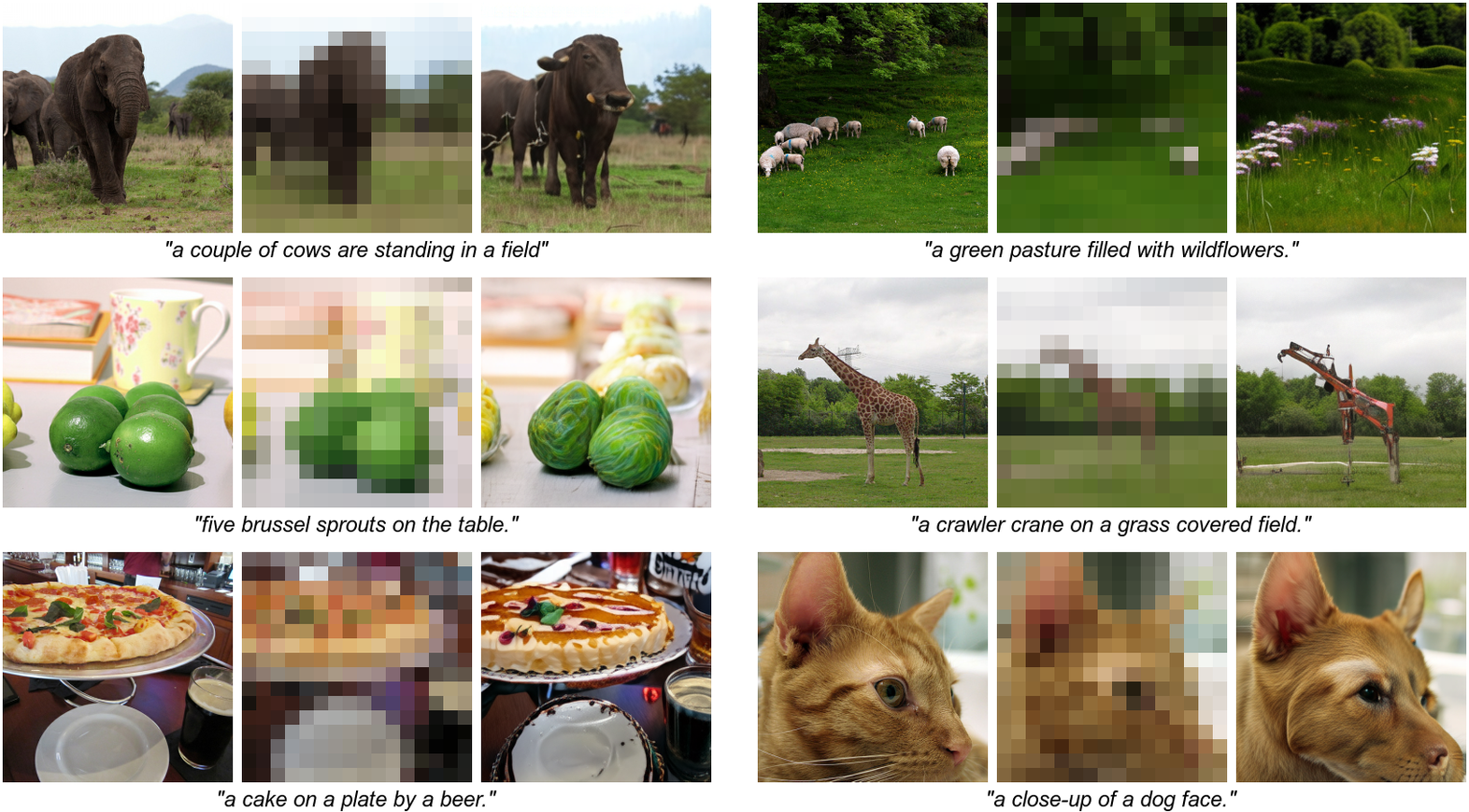}
   \caption{Text-guided image editing on MS-COCO~\cite{LinMBHPRDZ14}.
     Our coarse-to-fine generative models can do text-guided editing by encoding a 
     given image but keeping only the lower scales, and using a
     pre-trained AR-SIT to re-generate the higher scales conditioned on the textual prompt.
     Each triplet shows the given image, its $16\times 16$ reconstruction using only the coefficients
     used to start the generation, and the edited image after generating the whole sequence.
     \vspace{-0.5cm}
   }
   \label{fig:editing}
 \end{figure*}
 
\subsection{Text-guided image editing}
\label{sec:editing}
Our coarse-to-fine AR-SIT enables a text-guided image editing application,
where we want to change image details while keeping the same overall appearance,
which corresponds to freezing lower scales while generating higher.

We apply an AR-SIT trained only on the maximum likelihood objective as follows.
The given image is tokenized only up to the first scale,
corresponding to a coarse representation.
The tokenizer encoder must be scale-causal in order to avoid leakage from high to low scales, 
so we use SIT-SCE here. Now we use AR-SIT to generate the rest of the sequence, conditioning
on the textual caption.

In this experiment, we use a 5-scale model such that the lowest resolution is $16\times16$;
we found that starting with higher resolutions limits the changes the model
can generate. \cref{fig:editing} shows some examples of text-guided editing on
MS-COCO~\cite{LinMBHPRDZ14}, where we lightly modify the original captions,
for example by swapping ``elephants'' with ``cows''.
We show additional results in \cref*{fig:editing_supp} in the supplemental.

\begin{table}[t!]
  \begin{center}
    \resizebox{\columnwidth}{!}{    
      \begin{tabular}{
        @{}
        lccccc
        S[table-format=2.2,table-auto-round]
        S[table-format=3.1,table-auto-round]
        @{}}
      \toprule
                                              &  & params & seq len & CFG &  & {FID $\downarrow$} & {IS $\uparrow$} \\[0pt]
      \midrule
      AR-ViT-VQGAN (reported)                 &  & 650M   & 1024    & -   &  & 8.81      & 110.8    \\[0pt]
      AR-ViT-VQGAN (reproduced)               &  & 650M   & 1024    & -   &  & 8.37      & 111.76   \\[0pt]
      AR-SIT-4 (Ours)                         &  & 650M   & 1024    & -   &  & 6.95      & 138.27   \\[0pt]
      \midrule
      RQ-VAE~\cite{rqvae}                     &  & 480M   & 256     & -   &  & 15.72     & 86.8     \\[0pt]
      DQ-VAE~\cite{dqvae}                     &  & 355M   & 640     & -   &  & 7.34      & 152.8    \\[0pt]
      LlamaGen-L~\cite{sun2024autoregressive} &  & 343M   & 256     & 1.5 &  & 4.08      & 198.5    \\[0pt]
      VAR~\cite{VAR}                          &  & 310M   & 680     & 2.0 &  & 3.30      & 274.4    \\[0pt]
      AR-SIT-4* (Ours)                        &  & 350M   & 1024    & 1.5 &  & 4.06      & 190.9    \\[0pt]
      \midrule
    \end{tabular}
    }
  \end{center}
  \caption{ImageNet $256 \times 256$ class conditional generation.
    Top half shows a fair comparison against ViT-VQGAN, which we clearly outperform.
    AR-SIT-4* uses different hyperparameters for comparison against a broader
    class of methods.
    While we do not outperform all of them, there are differences in the architectures
    and training schedules that influence the results and are orthogonal to our contributions.
    Moreover, none of the baselines has the multiscale capabilities described in
    \ref{prop:tok_multiscale} and \ref{prop:ar_multiscale} (\cref{sec:intro}).
  }
  \label{tab:class_cond}
\end{table}

\subsection{Class-conditional image generation}
\label{sec:class_cond}
Here we evaluate class-conditional generation on $256 \times 256$ ImageNet.
We use the SIT-4 tokenizer from \cref{tab:multiscale_reconstruction}
and follow ViT-VQGAN's~\cite{yu2022vectorquantized} VIM-Base AR architecture
and training protocol. \cref{tab:class_cond} shows we outperform the baseline
in this fair comparison which is evidence for the improved conditioning
described in \ref{prop:ar_conditioning}.
\cref{fig:classcond_samples} shows samples.

AR-SIT-4* departs from ViT-VQGAN and Parti hyperparameters
to compare with more recent works,
by increasing the perceptual and adversarial loss weights,
doubling the tokenizer codebook, reducing its latent dimension,
training with a constant learning rate, including 2D RoPE~\cite{su2021roformer} and
GeGLU~\cite{shazeer2020gluvariantsimprovetransformer}, in both the tokenizer and generator.
We train the AR model longer and since there is no text encoder,
we double the number of decoder layers.
The bottom half of \cref{tab:class_cond} compares with similarly-sized AR models including some with
multiscale/residual latents
(although none exhibit properties \ref{prop:tok_multiscale} and \ref{prop:ar_multiscale}). 
AR-SIT-4* performance matches LlamaGen but not to VAR,
but the models are not exactly comparable since they use a convolutional tokenizers and
other improvements such as AdaLN~\cite{Peebles2022DiT} and attention normalization;
VAR also trains the tokenizer on larger data than ImageNet. 
These are orthogonal to our contributions and can be incorporated in future work.

We also train an AR-SIT-5* on $512 \times 512$ ImageNet and obtain an FID
of 5.97 and IS of 215.22, which shows the robustness of our
method at higher resolutions, while most prior work requires separate
stages for upsampling.
\cref{fig:classcond512_samples} shows samples.

\section{Conclusion and limitations}
\label{sec:conclusion}
We presented a spectral image tokenizer (SIT),
and an autoregressive generative transformer trained with it (AR-SIT).
SIT is naturally multiscale and leverages spectral properties of natural images
for improved reconstruction quality.
AR-SIT enables applications such as rapid generation
of coarse images that can be refined later, and text-guided image upsampling and editing.

While our methods improved tokenizer reconstruction accuracy and 
class-conditional generation, the text-to-image metrics were similar to the fair baseline.
\citet{pmlr-v202-chang23b} similarly observed that a better tokenizer does not necessarily
lead to a better generative model.
Nevertheless, our method has multiscale properties and enables new applications
not possible with prior work. 
We only experiment with realtively small AR-SITs of 350-650M parameters, while
the Parti~\cite{yu2022scaling} baseline goes up to 22B parameters.

\section{Acknowledgments}
We thank Leonardo Zepeda-Núñez for reviewing this manuscript and
offering interesting discussions and suggestions, and Jon Barron for
sharing useful code we relied on.

{
    \small
    \bibliographystyle{ieeenat_fullname}
    \bibliography{main}
}

\clearpage
\setcounter{page}{1}
\appendix 
\maketitlesupplementary

\section{Ablation study}
We conduct an ablation study to evaluate the effects of our design decisions.
Starting from the model denoted ``SIT-4'' in \cref{tab:multiscale_reconstruction},
we train SIT using various number of scales, sequences lengths, vocabulary sizes, wavelet families,
and using scale-causal attention on the encoder and/or decoder.
\cref{tab:ablation} shows the results.

We note that while increasing the vocabulary size or sequence length improves reconstruction,
it is at the cost of making the task harder for generation so these tokenizers did not actually
lead to better generative models. This has also been observed by prior and concurrent work \cite{pmlr-v202-chang23b,ramanujan2024worsebetternavigatingcompressiongeneration,yao2025vavae,hansenestruch2025learningsscalingvisualtokenizers}.

Another interest result is that the simple Haar wavelet performs better than the ones that are
widely used for compression: LeGall5/3 and CDF9/7.
We hypothesize it's related to the larger filter supports, since our results show
worse reconstruction the larger the support.
It could be due to increased padding and leakage of information between neighboring patches;
the same trend is observed for both reflective and zero padding.
Some related work with wavelets in diffusion models and GANs also use Haar \cite{swagan,phung2023wavediff},
but they did not report ablations. 

\section{Implementation details}
\subsection{Multiscale reconstruction}
\label{supp:multiscale_reconstruction}
For the multiscale image reconstruction experiments in \cref{sec:multiscale_reconstruction},
we train SIT following ViT-VQGAN~\cite{yu2022vectorquantized} closely.
We use the ``small'' encoder and ``small'' decoder as described in
ViT-VQGAN~\cite{yu2022vectorquantized}.
Each transformer layer comprises a layer norm, self-attention, residual connection, followed by
an MLP with layer norm, two dense layers, and a second residual connection.
The ``small'' configuration has 8 layers, 8 heads, feature dimension of 512 that
increases to 2048 in the MLP hidden layer (the hidden dimension), then back to 512.
Learnable positional embedding is added to the transformer input.
The codebook size is \num{8192} for both codebooks (approximation and details).

The tokenizer is trained for \num{500000} steps with a learning rate that linearly
increases up to \num{1e-4} during the first \si{10}{\%} steps, then decays
following a cosine schedule.
We use batch size 256 and L2 weight decay with a \num{1e-4} factor.
The loss components are weighted as follows: 1.0 for L2, 0.1 for perceptual, 0.1 for adversarial,
and \num{0.25} for codebook commitment.

\subsection{Coarse-to-fine text-to-image generation}
\label{supp:t2i}
For the text-to-image generation experiments in \cref{sec:partial_generation}, we train SIT models
with a ``small'' encoder and ``large'' decoder as defined in ViT-VQGAN~\cite{yu2022vectorquantized}
and used by Parti~\cite{yu2022scaling}. The ``small'' transformer has 8 layers with 8 heads,
feature dimension 512 and hidden dimension 2048.
The ``large'' transformer has 32 layers with 16 heads,
feature dimension 1280 and hidden dimension 5120.
Differently from Parti, we train this tokenizer once from scratch instead of in 
two stages; Parti first trains a ``small'' encoder and decoder, then freezes the ``small''
encoder and trains ``large'' decoder.

For the generative AR-SIT, we follow the smallest architecture presented in Parti~\cite{yu2022scaling},
with 12 layers in the text encoder and 12 decoder layers, 16 heads, 
1024 feature dimensions and 4096 hidden dimensions.
The text conditioning is through cross-attention.
We use classifier-free guidance (CFG)~\cite{ho2022classifierfreediffusionguidance} scale of 3.0.
No reranking is used.

AR-SIT is trained for \num{500000} steps with a learning rate that linearly
increases up to \num{4.5e-4} during the first \si{10}{\%} steps,
then decays exponentially.
We use batch size 256 and the loss is the just the softmax cross-entropy.

\subsection{Class-conditional generation}
For the class-conditional image generation fair comparison experiments in \cref{sec:class_cond},
we again follow ViT-VQGAN closely
and train SIT-4 with ``small'' encoder and ``small'' decoder as defined in
\cref{supp:multiscale_reconstruction}, and AR-SIT-4 is based on VIM-Base with 24 transformer layers,
16 heads, 1536 model dimension, 6144 hidden dimension, and dropout ratio of 0.1.
The model is trained for \num{500000} steps with a learning rate that linearly
increases up to \num{4.5e-4} during the first \si{10}{\%} steps,
then decays exponentially. We use batch size 1024. No CFG or reranking is used.

AR-SIT-4* modifies both the tokenizer and autoregressive models and training.
The main architecture improvements are the use of GeGLU activations on all MLPs
and axial 2D RoPE for both the tokenizer and AR model.
The tokenizer has a latent dimension of 8,
vocabulary size of \num{16384} for the approximation and details codebooks,
is trained for \num{300000} steps with a batch size of 128, constant
learning rate of \num{1e-4}, L2 weight decay of 0.05, and loss weights
of 1.0 for L2, 1.0 for LPIPS, 0.5 for adversarial, \num{0.25} for codebook commitment.
The adversarial training starts at \num{20000} steps.
The AR model has 24 transformer layers,
16 heads, 1024 model dimension, 4096 hidden dimension,
dropout ratio of 0.1, and is trained for 1.2M steps with batch size 256
and learning rate of \num{1e-4}. We weight the loss such that each scale has a
weight 4x smaller than the previous and found that it improves performance slightly.

\subsection{Metrics}
Throughout our experiments we report the following metrics:
\begin{itemize}
\item FID - Fréchet Inception Distance~\cite{fid}
  estimates the distance between the distribution of generated and ground truth features obtained from a pre-trained inception model, by assuming such
  distributions are Gaussians and applying the Fréchet distance.
\item IS - Inception Score (IS)~\cite{inceptionscore}
  measures the entropy of a pre-trained classifier
  on the generated images, where low entropy is expected for good quality images.
\item LPIPS - Learned Perceptual Image Patch Similarity~\cite{lpips}
  measures the distance between
  visual features reconstructed and ground truth images, where the features come from a pre-trained
  model.
\item PSNR - Peak Signal-to-Noise Ratio is a pixel-wise similarity metric, the negative log of the mean-squared error.
\end{itemize}

FID and IS evaluate the quality of generated images without a corresponding ground truth,
while LPIPS and PSNR are used when we have a ground truth, for example when evaluating the tokenizer.

\begin{table}[t!]
  \begin{center}
      \begin{tabular}{cl
        S[table-format=1.2,table-auto-round]
        S[table-format=2.2,table-auto-round]
        S[table-format=2.2,table-auto-round]
        }
        \toprule
        Input resolution && {FID}  & {IS}    & {PSNR}  \\[0pt]
        \midrule
        $32\times32$           && 6.19 & 32.46 & 19.05 \\[0pt]
        $16\times16$           && 7.53 & 31.75 & 16.74 \\[0pt]
        \bottomrule
      \end{tabular}
    \end{center}
  \caption{Image upsampling metrics on MS-COCO~\cite{LinMBHPRDZ14}.}
  \label{tab:upsampling}    
  \end{table}

\begin{table*}[t]
\begin{center}
  \begin{tabular}
    {lccccccc
    S[table-format=1.3,table-auto-round]
    S[table-format=2.2,table-auto-round]
    S[table-format=1.2,table-auto-round]
    S[table-format=3.1,table-auto-round]
    }
\toprule
model     & num scales & seq len & vocab & wavelet   & SCE    & SCD    && {LPIPS $\downarrow$} & {PSNR $\uparrow$}  & {FID $\downarrow$}  & {IS $\uparrow$}    \\[0pt]
\midrule
ViT-VQGAN & -         & 1024    & 8192  & -         & -      & -      && 0.164 & 23.76 & 1.20 & 194.6 \\[0pt]
SIT-4     & 4         & 1024    & 8192  & Haar      & \xmark & \xmark && 0.143 & 24.01 & 1.20 & 199.5 \\[0pt]
SIT-5     & 5         & 1280    & 8192  & Haar      & \xmark & \xmark && 0.135 & 24.48 & 0.97 & 202.3 \\[0pt]
\midrule
SIT-4     & 4         & 1024    & 8192  & Haar      & \xmark & \(\checkmark\)  && 0.166 & 23.51 & 1.45 & 191.3 \\[0pt]
SIT-4     & 4         & 1024    & 8192  & Haar      & \(\checkmark\)  & \(\checkmark\)  && 0.184 & 23.41 & 1.97 & 179.8 \\[0pt]
\midrule
SIT-5     & 5         & 1280    & 8192  & LeGall5/3 & \xmark & \xmark && 0.145 & 24.35 & 1.10 & 198.0 \\[0pt]
SIT-5     & 5         & 1280    & 8192  & CDF9/7    & \xmark & \xmark && 0.147 & 24.31 & 1.14 & 198.1 \\[0pt]
\midrule
SIT-5     & 5         & 1280    & 4096  & Haar      & \xmark & \xmark && 0.155 & 23.66 & 1.33 & 193.9 \\[0pt]
SIT-5     & 5         & 1280    & 16384 & Haar      & \xmark & \xmark && 0.134 & 24.62 & 0.91 & 203.0 \\[0pt]
\midrule
SIT-2     & 2         & 2048    & 8192  & Haar      & \xmark & \xmark && 0.097 & 26.21 & 0.77 & 212.0 \\[0pt]
SIT-3     & 3         & 768     & 8192  & Haar      & \xmark & \xmark && 0.186 & 23.12 & 2.04 & 176.8 \\[0pt]
SIT-2     & 2         & 512     & 8192  & Haar      & \xmark & \xmark && 0.244 & 20.93 & 4.30 & 141.2 \\[0pt]
SIT-6     & 6         & 384     & 8192  & Haar      & \xmark & \xmark && 0.273 & 20.72 & 6.25 & 129.9 \\[0pt]
SIT-5     & 5         & 320     & 8192  & Haar      & \xmark & \xmark && 0.283 & 20.51 & 7.76 & 118.0 \\[0pt]
\bottomrule
\end{tabular}
\end{center}
  \caption{Ablation study. We report ImageNet reconstruction metrics for SIT variations,
    following the ViT-VQGAN protocol from \cref{tab:multiscale_reconstruction}.
    We evaluate the effect of number of scales, sequence length, vocabulary size, and wavelet family.
    We also evaluate the scale-causal attention on the encoder (SCE) and decoder (SCD),
    while it generally reduces reconstruction accuracy, it enables
    the various multiscale properties demonstrated in the text.
  }
  \label{tab:ablation}
\end{table*}

\section{Additional results}
\subsection{Multiscale reconstruction}
\begin{figure*}[t]
  \centering
   \includegraphics[width=\linewidth]{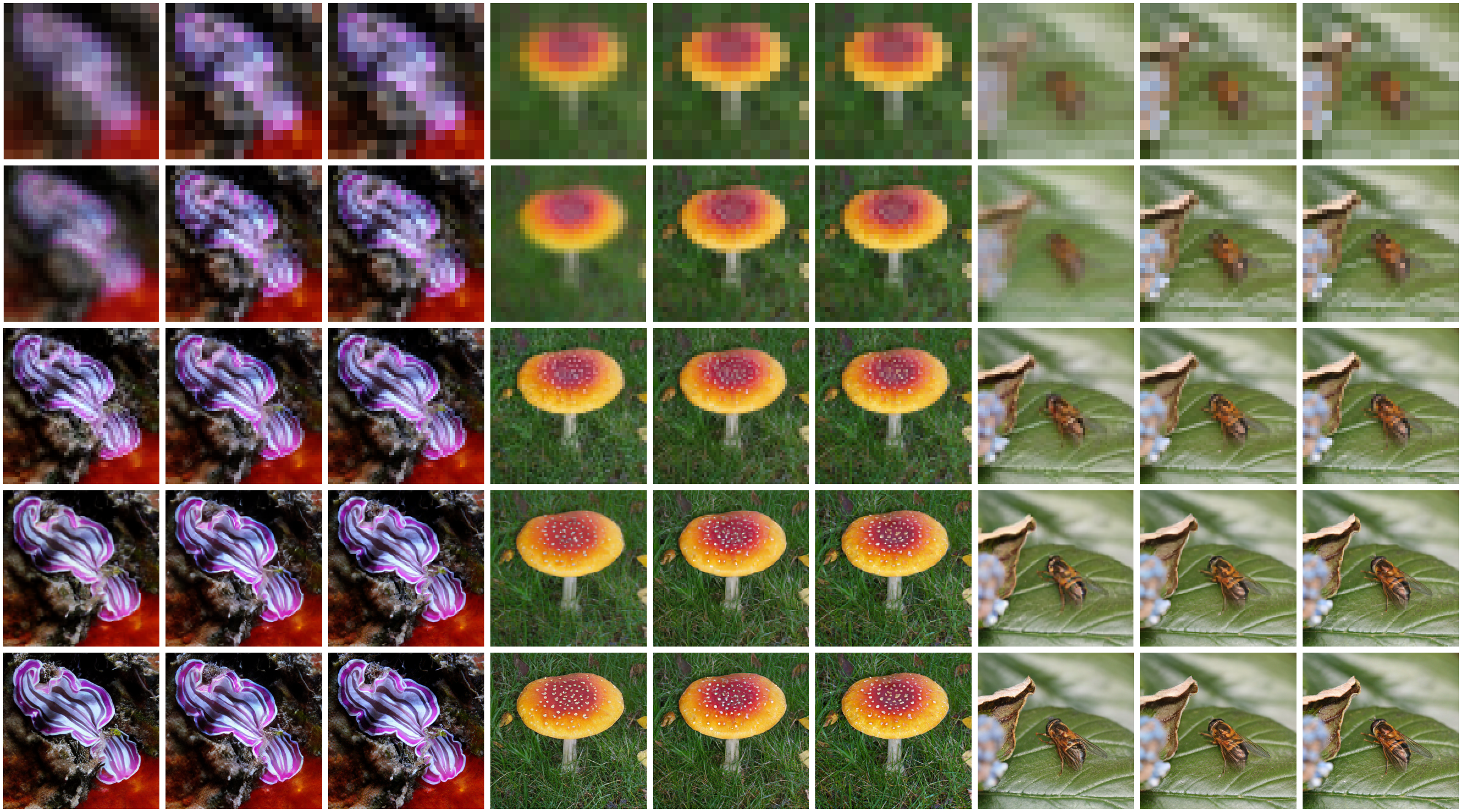}
   \caption{Multiscale reconstruction on ImageNet.
     Each triplet shows reconstruction from the ViT-VQGAN baseline~\cite{yu2022vectorquantized},
     our SIT-SC-5 (Spectral Image Tokenizer with Scale-Causal attention and 5 scales),
     and the ground truth.
     Each row shows $4\times$ as many pixel inputs as the previous,
     with the first row corresponding to $16\times16$ resolution, and the last to $256\times256$.
     Our method is naturally multiresolution, significantly outperforming
     the baseline on lower resolutions even when trained only on $256\times256$ inputs,
     while achieving similar accuracy on higher resolutions.
}
   \label{fig:multiscale_reconstruction}
\end{figure*}

\cref{fig:multiscale_reconstruction} shows reconstruction samples at multiple resolutions,
for the experiments described in \cref{sec:multiscale_reconstruction}.

\subsection{Text-guided upsampling}
\begin{figure*}[h!]
  \centering
   \includegraphics[width=\linewidth]{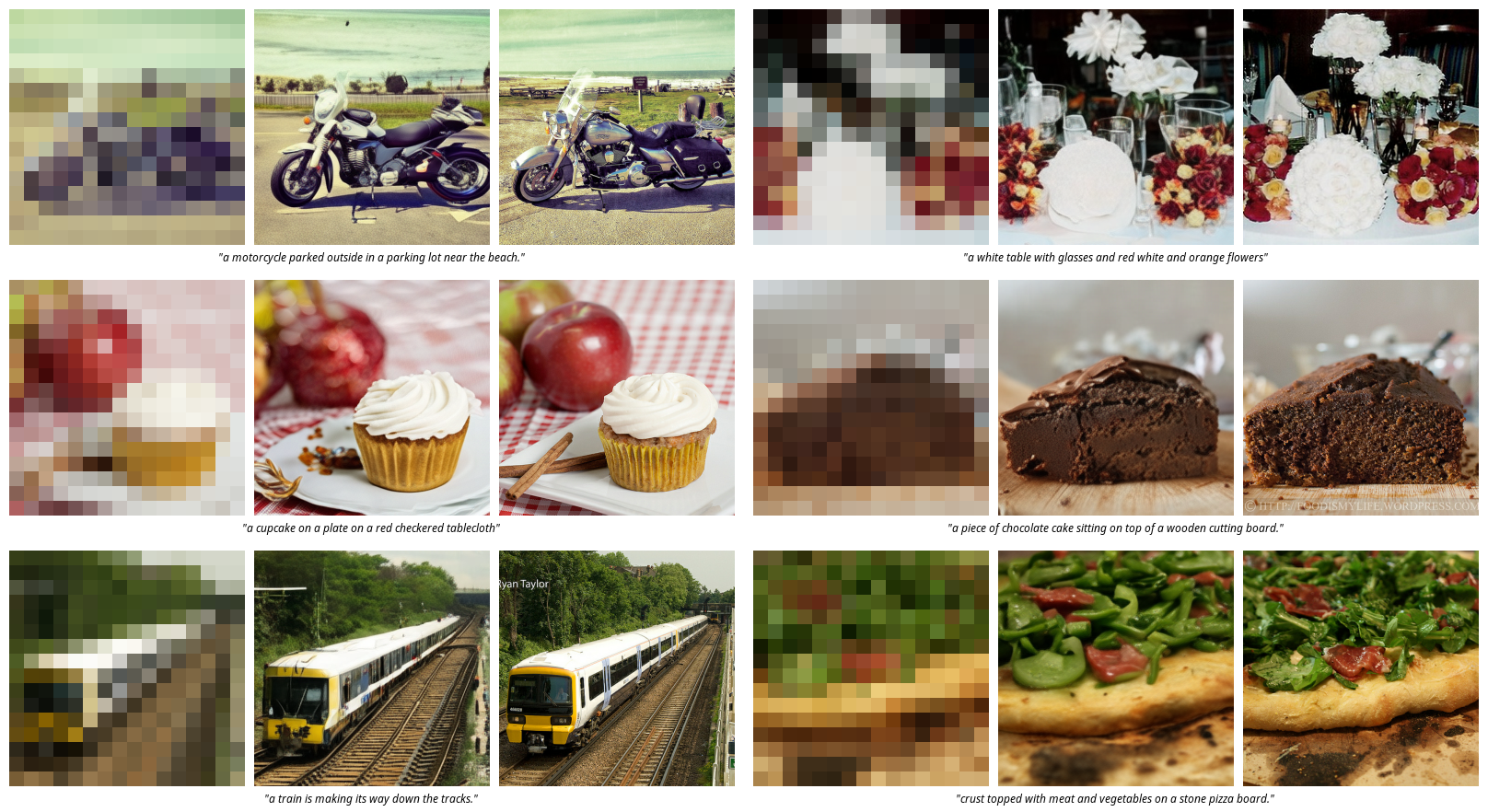}
   \caption{Additional text-guided image upsampling results.
     Here we consider the more challenging task of upsampling from $16\times16$ to $256\times256$.
     Each triplet shows the given $16\times16$ image, our $256\times256$ reconstruction and the ground truth.
   }
   \label{fig:upsampling_supp}
 \end{figure*}

In this section, we show additional results to the text-guided image upsampling experiments
from \cref{sec:upsampling}. We evaluate the same task on the more challenging case where
we upsample a $16 \times 16$ input to $256 \times 256$. \cref{tab:upsampling} shows the metrics
and \cref{fig:upsampling_supp} shows some examples.

\begin{figure*}[t!]
  \centering
   \includegraphics[width=\linewidth]{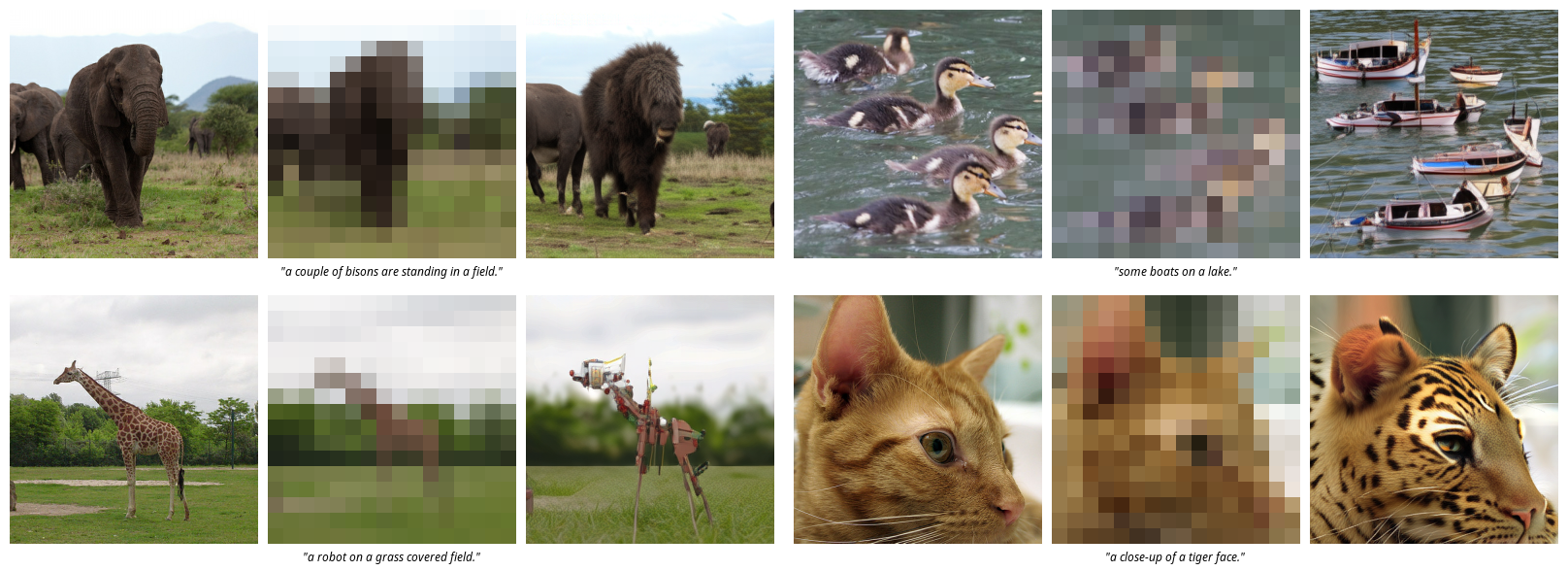}
   \caption{Additional results for text-guided image editing on MS-COCO~\cite{LinMBHPRDZ14}.
     Each triplet shows the given image, its reconstruction given only the coefficients
     used to start the generation, and the edited image after generating the whole sequence.
     The guiding prompt is shown under each triplet.
   }
   \label{fig:editing_supp}
 \end{figure*}
 
\subsection{Additional editing results}
\cref{fig:editing_supp} shows additional results for the text-guided image editing experiment
described in \cref{sec:editing}.

\subsection{Class-conditional generated samples}
\begin{figure*}[t!]
  \centering
   \includegraphics[width=\linewidth]{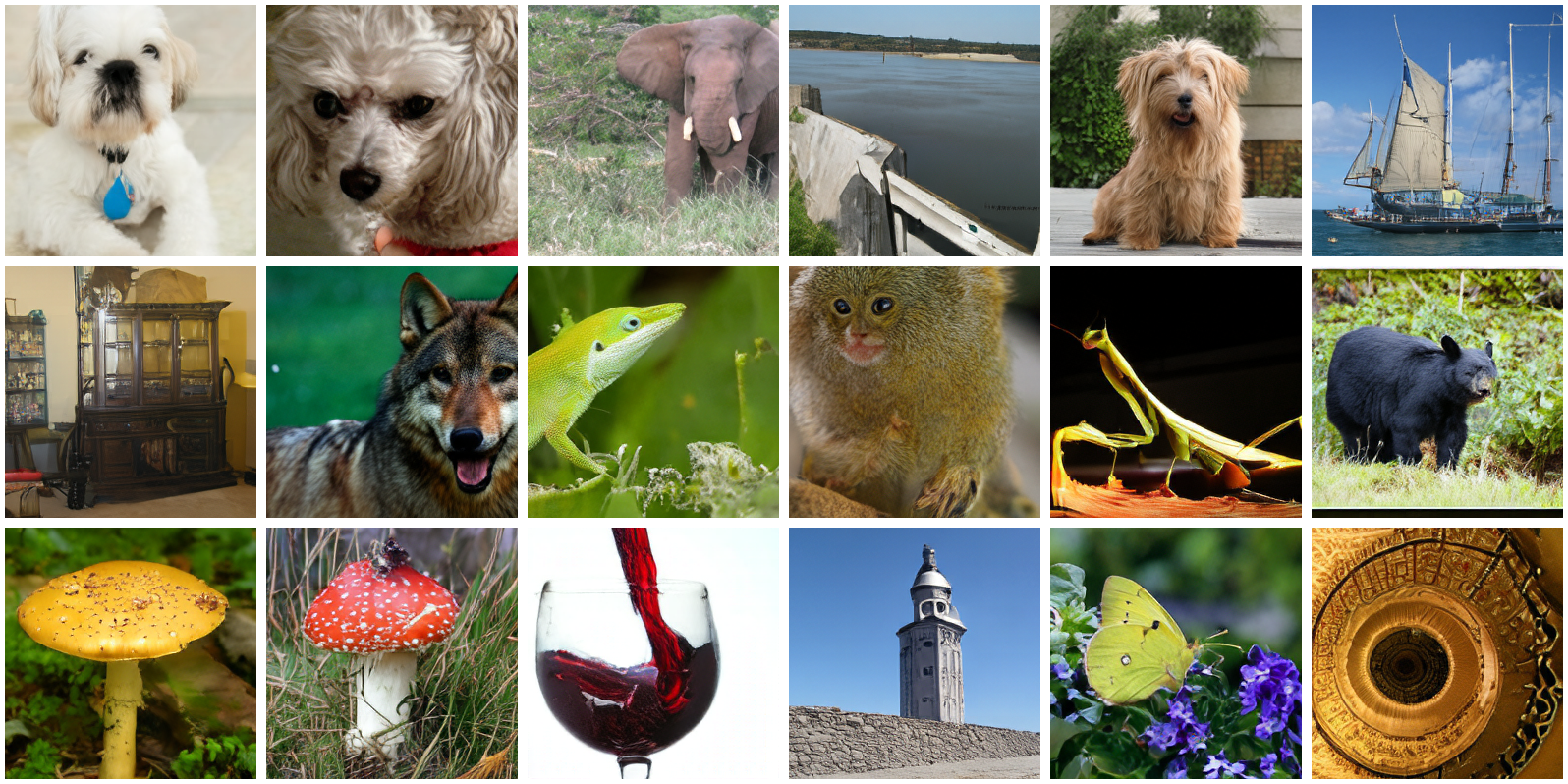}
   \caption{Samples of class-conditional generation with AR-SIT-4 on $256 \times 256$ ImageNet.
   }
   \label{fig:classcond_samples}
\end{figure*}

\begin{figure*}[t!]
  \centering
   \includegraphics[width=\linewidth]{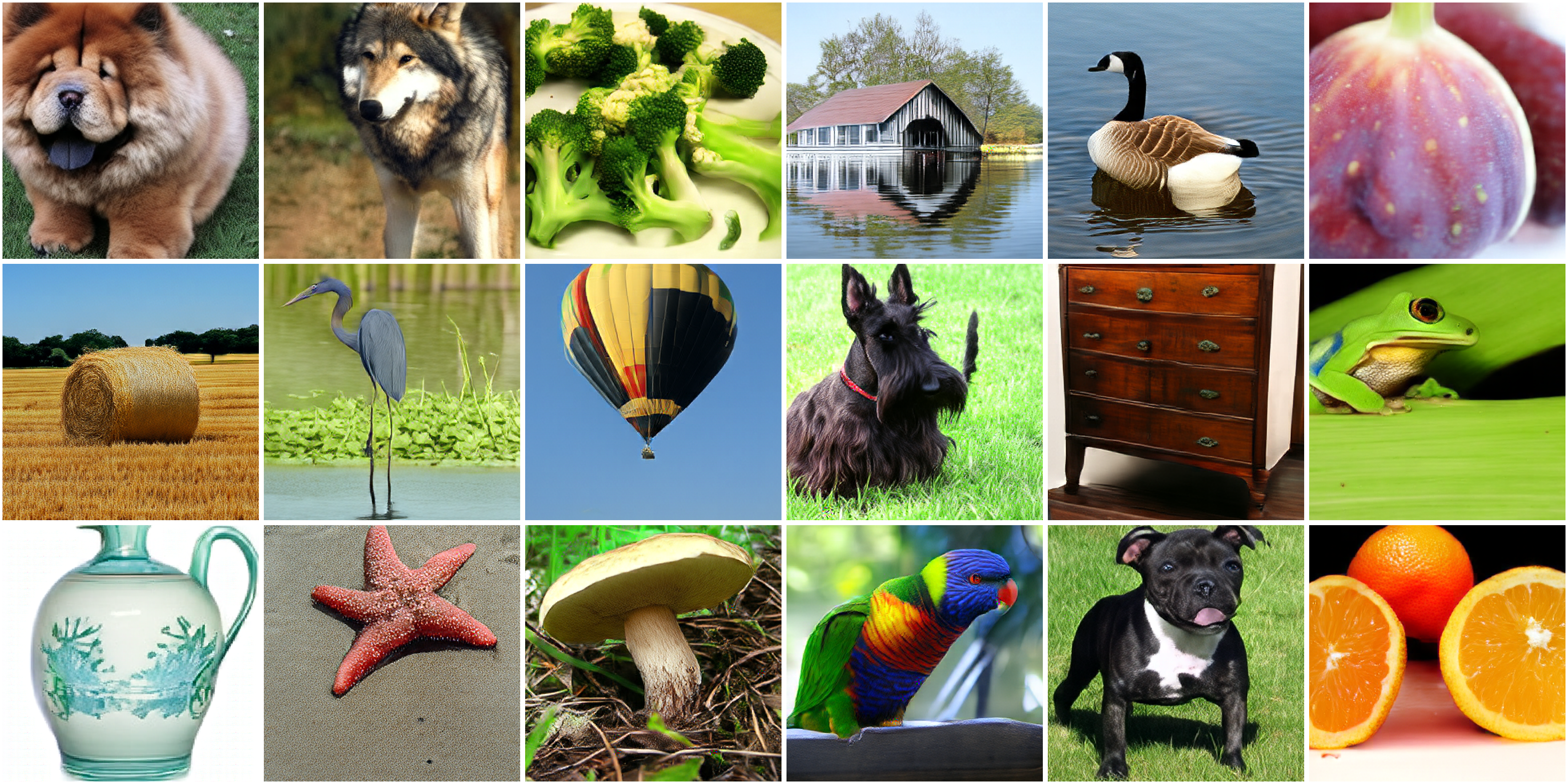}
   \caption{Samples of class-conditional generation with AR-SIT-5* on $512 \times 512$ ImageNet.
   }
   \label{fig:classcond512_samples}
\end{figure*}

\cref{fig:classcond_samples} shows samples of class-conditional $256 \times 256$ ImageNet generation, and \cref{fig:classcond512_samples} shows $512 \times 512$ ImageNet generations
as described in \cref{sec:class_cond}.

\end{document}